\documentclass[11pt]{article}

\usepackage{acl}
\usepackage{enumitem}
\usepackage{times}
\usepackage{latexsym}

\usepackage[T1]{fontenc}

\usepackage[utf8]{inputenc}

\usepackage{microtype}

\usepackage{mathrsfs}
\usepackage{amsmath}
\usepackage{amssymb}
\usepackage{graphicx}
\usepackage{subcaption}
\usepackage{multirow} 
\usepackage{array} 
\usepackage{booktabs} 
\usepackage{siunitx} 
\usepackage{makecell} 
\usepackage{adjustbox} 
\usepackage{amssymb} 
\usepackage{hyperref}
\usepackage{xcolor}     
\usepackage{url}        
\usepackage{geometry}
\usepackage{tabularx}
\usepackage[most]{tcolorbox} 
\usepackage{listings}
\usepackage[affil-it]{authblk}

\hypersetup{
    colorlinks=true,
    linkcolor=blue,
    filecolor=magenta,      
    urlcolor=blue,
    pdftitle={Overleaf Example},
    pdfpagemode=FullScreen,
}
\definecolor{hallucination_red}{RGB}{200, 0, 0}
\definecolor{incorrect_orange}{RGB}{200, 100, 0}
\definecolor{missing_purple}{RGB}{100, 0, 150}
\definecolor{imprecise_blue}{RGB}{0, 100, 150}

\definecolor{maincolor}{RGB}{45, 52, 54}  
\definecolor{backcolor}{RGB}{250, 250, 250} 
\definecolor{keycolor}{RGB}{178, 34, 34} 
\definecolor{stringcolor}{RGB}{0, 100, 0} 

\lstdefinestyle{promptstyle}{
    basicstyle=\ttfamily\footnotesize, 
    basewidth={0.5em, 0.45em},       
    columns=flexible,                
    keepspaces=true,                 
    breaklines=true,                 
    breakindent=1em,                 
    keywordstyle=\color{blue}\bfseries, 
    stringstyle=\color{violet},      
    commentstyle=\color{gray}\itshape,  
    frame=none,                      
    aboveskip=0pt,                   
    belowskip=0pt,                   
    xleftmargin=0pt,                 
    xrightmargin=0pt,                
    escapeinside={(*@}{@*)},         
    morekeywords={score, reason, decision, target_count, entity_type, action, element_id}
}

\title{See and Remember: A Multimodal Agent for Web Traversal}

\author{
Xinjun Wang$^{1}$, Shengyao Wang$^{1}$, Aimin Zhou$^{1, 2}$, and \textbf{Hao Hao}$^{1, }$ \thanks{~~Corresponding author: hhao@mail.ecnu.edu.cn} \\
\\
$^{1}$~Shanghai Institute of AI for Education, East China Normal University \\
$^{2}$~Shanghai Innovation Institute
}

\begin{document}
\maketitle
\begin{abstract}
Autonomous web navigation requires agents to perceive complex visual environments and maintain long-term context, yet current Large Language Model (LLM) based agents often struggle with spatial disorientation and navigation loops. In this paper, we propose generally applicable V-GEMS~(Visual Grounding and Explicit Memory System), a robust multimodal agent architecture designed for precise and resilient web traversal. Our agent integrates visual grounding to resolve ambiguous interactive elements and introduces an explicit memory stack with state tracking. This dual mechanism allows the agent to maintain a structured map of its traversal path, enabling valid backtracking and preventing cyclical failures in deep navigation tasks. We also introduce an updatable dynamic benchmark to rigorously evaluate adaptability. Experiments show V-GEMS significantly dominates the WebWalker baseline, achieving a substantial 28.7\% performance gain. Code is available at \href{https://github.com/Vaultttttttttttt/V-GEMS}{V-GEMS}.
\end{abstract}

\section{Introduction}
\label{sec:intro}

The World Wide Web serves as the largest dynamic repository of human knowledge. Consequently, autonomous web navigation has emerged as a paramount capability for Artificial Intelligence, fueling advancements in Large Language Model (LLM) training~\citep{dong2019unified, guu2020retrieval, hoffmann2022empirical}, real-time Deep Research,~\citep{shao-etal-2024-assisting, openai2025deepresearch} and open-ended workflow automation. Moving beyond static scraping scripts, the field is undergoing a paradigm shift toward intelligent agents capable of interacting with dynamic Graphical User Interfaces (GUIs) directly, mirroring human browsing behavior~\citep{nguyen2025gui}.

In this evolution, Vision-Language Models (VLMs) have become indispensable~\citep{zhang2024vision}. Unlike traditional methods that rely solely on raw HTML Document Object Models (DOM), VLMs empower agents with the ability to perceive layout semantics, interpret icons, and interact with complex UI elements that are often obfuscated in text-only representations~\citep{yang2024embodied}. This visual modality is critical for bridging the gap between abstract user instructions and concrete executable actions on modern, media-rich websites.

Recent research has made significant strides in this direction. Notably, \textbf{WebWalker}~\citep{wu2025webwalker} established a robust baseline by introducing a dual-agent (Explorer-Critic) architecture, demonstrating impressive capabilities in task decomposition and correctness verification. This pioneering framework highlights the potential of LLM-based planning. However, relying predominantly on textual DOM parsing or standard LLM reasoning leaves room for optimization in highly complex environments~\citep{tang2025survey}. For instance, without the pixel-level intuition provided by VLMs, agents may face challenges in disambiguating visually dense elements, leading to suboptimal click operations. Furthermore, the inherent structural complexity of the web characterized by deep hierarchies and dead-end links can occasionally cause stochastic agents to experience ``navigational disorientation'', where they struggle to maintain context over long-horizon trajectories~\citep{koh2024visualwebarena}.

Despite these advancements, developing a truly resilient autonomous agent remains a formidable challenge. Current navigation paradigms generally encounter three systemic bottlenecks:
\begin{itemize}
    \item \textbf{Modal Inefficiency:} Text-based DOMs often fail to capture the picture of a page (e.g., distinguishing a clickable banner from an image). Relying solely on text can lead to missed interaction opportunities or hallucinations about page content.
    \item \textbf{Navigational Stagnation:} In the absence of explicit state tracking, agents lack a cognitive map. They prone to getting stuck in loops or failing to backtrack correctly after exploring deep sub-pages, behaving like a browser without a ``Back'' button.
    \item \textbf{Symbolic Inaccuracy:} General-purpose LLMs frequently exhibit deficits in precise quantitative reasoning tasks (e.g., Find exactly 5 papers)~\citep{bubeck2023sparks}, leading to aggregation errors even when the correct information is found.
\end{itemize}

To surmount these challenges, we propose \textbf{V-GEMS} (\textbf{V}isual-\textbf{G}rounded \textbf{E}xploration with \textbf{M}emory and \textbf{S}ymbolic tools). Building upon the strengths of dual-agent frameworks, V-GEMS augments the navigation process with a synergistic toolkit comprising three specialized modules designed to enhance robustness and precision. 
First, we introduce an Adaptive US~(Understanding Score) Calculator that dynamically integrates VLM perception to resolve visual ambiguity only when necessary. 
Second, we implement a Stateful URL Stack, a deterministic memory structure that provides the agent with topological awareness, enabling precise backtracking and pruning of visited paths. 
Third, we deploy a Symbolic Counter to decouple arithmetic operations from the LLM, ensuring high-fidelity data aggregation.

Furthermore, recognizing that static benchmarks rapidly become obsolete in the live web, we introduce an updatable evaluation pipeline. This ensures our agent is tested against the temporal reality of the internet.

Our main contributions are summarized as follows:
\begin{itemize}
    \item We propose V-GEMS, a novel framework that integrates adaptive visual perception with explicit memory and symbolic tools. This architecture effectively resolves the navigational problems and arithmetic hallucinations common in previous approaches.
    \item We construct a reproducible and extensible evaluation pipeline that generates Q\&A tasks from live websites, addressing the issue of benchmark obsolescence and ensuring fair, real-time assessment.
    \item Extensive evaluations demonstrate that V-GEMS achieves a significant performance improvement over state-of-the-art baselines, validating the effectiveness of our modular toolkit in complex, domain-specific retrieval tasks.
\end{itemize}

\section{Related Work}
\label{sec:related}

\subsection{Web Automation and Multimodality}
The initial efforts in interacting with the web focused heavily on rule-based or behavioral cloning techniques, primarily aiming for reliable Web Automation~\citep{liu2018reinforcement, humphreys2022data}. These systems relied on structured DOM parsing and rule-based path selectors (such as XPath) to execute predefined tasks, often struggling with generalization or handling websites that lacked clean structural annotations~\citep{sager2025comprehensive}. With the rise of deep learning, this field shifted towards leveraging language models. Recently, the integration of Multimodality has emerged as a crucial solution to address visual ambiguities~\citep{hong2024cogagent,cheng2024seeclick}. Agents now frequently incorporate VLMs to process visual elements (e.g., hidden navigation elements, image-based buttons)~\citep{he-etal-2024-webvoyager, koh2024visualwebarena}, moving beyond the limitations of text-only DOM parsing. Our work builds on this multimodal foundation, but further focuses on optimizing the agent's internal cognitive process to maximize both efficiency and reliability through the switch between LLM and VLM specifically.

\subsection{Web Navigation Agents}
Existing Web Navigation Agents can generally be categorized into two lines of research. The first emphasizes training smaller, specialized models to enhance specific aspects of navigation, such as filtering viable actions or identifying key HTML elements~\citep{seeact,mt_mind2web_agent,furuta2024multimodal}. The second, and more dominant approach, utilizes the strong reasoning capabilities of Large Language Models (LLMs) within structured frameworks like ReAct~\citep{react, xu2025towards, lu2023chameleon}. These models excel at task planning but frequently suffer from two critical cognitive deficits. First, they exhibit \textit{arithmetic hallucination} when quantitative tasks (e.g., counting instances) are required. Second, they lack explicit state management, leading to \textit{navigational amnesia} or getting stuck in recursive website hierarchies. Our proposed agent directly addresses these shortcomings by introducing an explicit symbolic counter and a stateful url stack, transforming the agent from a purely reactive planner to a memory-augmented, reliable explorer.

\subsection{Web Traversal Benchmarks}
The progression of Web Traversal Benchmarks has paralleled agent development, moving from simple web-scraping scenarios to complex instruction-following tasks~\citep{liu2018reinforcement,xu-etal-2021-grounding,humphreys2022data,yao2022webshop,mialon2024gaia,xu2024turkingbenchchallengebenchmarkweb}. Mind2Web~\citep{deng2023mindweb} established the standard for evaluating complex web interaction via multiple-choice questions. More recent benchmarks, such as MMInA~\citep{zhang2024mmina} and AssistantBench~\citep{assistantbench}, have focused on challenging, time-consuming tasks requiring multi-page navigation, demonstrating the increasing complexity sought by the community. However, a major hurdle remains: the Temporal Validity Gap. Since the real-world web is dynamic, most static evaluation datasets rapidly become outdated, severely limiting experimental reproducibility~\citep{hantke2025web, rauber2021precisely}. Unlike prior works that rely on fixed snapshots, our research proposes an Updatable and Extensible Evaluation Pipeline, ensuring that the resulting EverWebQA benchmark remains current and relevant for validating web agent capabilities over time.

\section{Proposed Method}
\label{sec:proposed method}

\begin{figure*}[t]
  \centering
  \includegraphics[width=\textwidth]{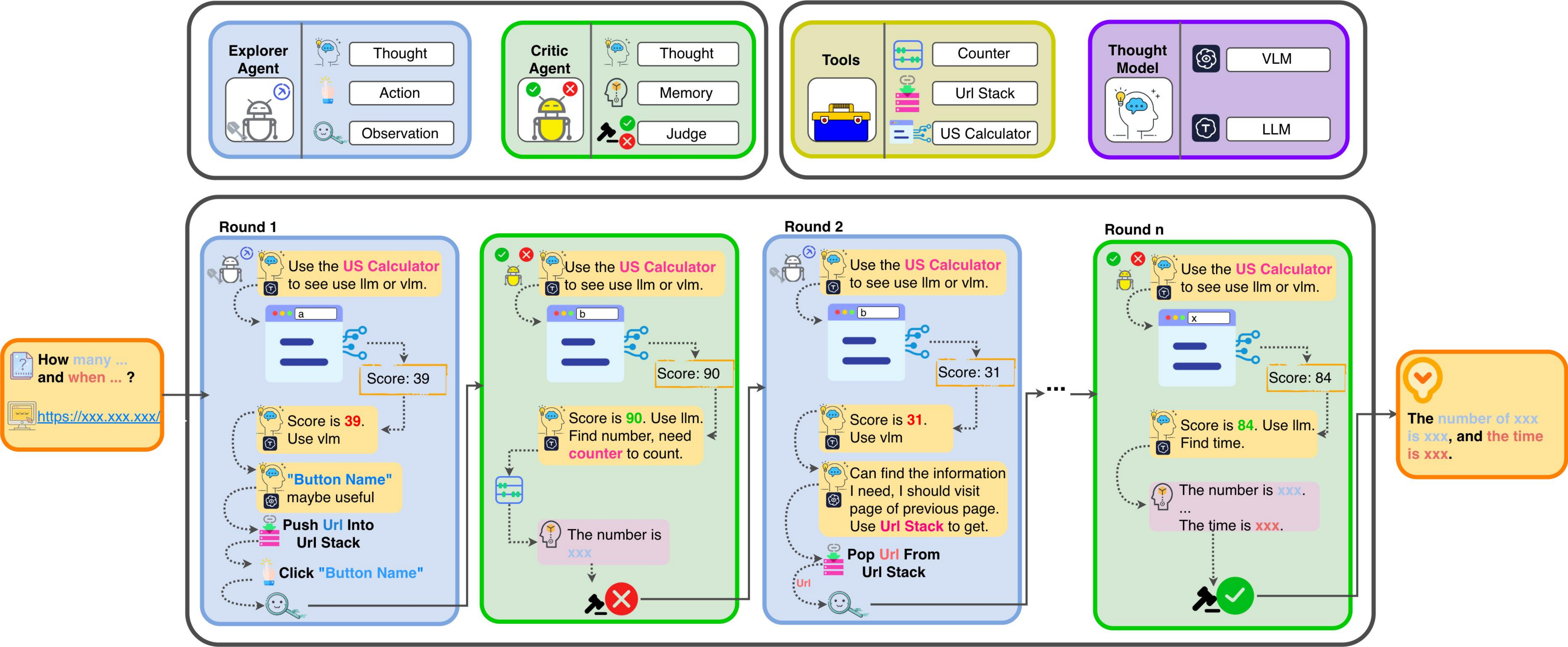}
  \caption{\textbf{The framework of V-GEMS.} Our system augments the dual-agent architecture (Explorer and Critic)  with a specialized \textbf{Tools} suite consisting of a Counter, URL Stack, and US Calculator. The \textbf{US Calculator} adaptively scores page content to decide between LLM and VLM processing. The \textbf{URL Stack} enables stateful backtracking, while the \textbf{Counter} ensures arithmetic precision across multiple navigation steps.\textit{For a comprehensive visualization of the end-to-end execution workflow, please refer to Appendix~\ref{sec:demonstration}.}}
  \label{fig:framework}
\end{figure*}

Based on the dual-agent scaffolding of WebWalker~\citep{wu2025webwalker}, we introduce V-GEMS, a system equipped with a modular toolkit to resolve bottlenecks in memory, calculation, and perception. As shown in Figure \ref{fig:framework}, our method consists of three key components. First, to mitigate arithmetic hallucinations, we deploy a Symbolic Counter that ensures deterministic tracking of entities. Second, to prevent navigational stagnation, we implement a URL Stack that enables stateful backtracking in deep hierarchies. Finally, to balance cost and performance, we utilize a US Calculator to adaptively trigger VLM assistance only when textual representations are insufficient.

\subsection{Symbolic Counter with Adaptive Termination}
\label{sec:counter}

A pervasive challenge in autonomous agents is arithmetic hallucinations~\citep{bubeck2023sparks}, where LLMs fail to maintain arithmetic consistency across long-horizon trajectories or struggle to determine optimal stopping conditions. As illustrated in the \textbf{Tools} module of Figure \ref{fig:framework}, we decouple quantitative tracking from linguistic reasoning by introducing an external \textbf{Counter Module} acting as a deterministic logic guardrail.

Unlike passive accumulators, our Counter functions as a programmable controller that parses constraints from the user query to activate one of two distinct execution modes (comprehensive logic flow detailed in \textbf{Appendix \ref{sec:appendix_counter_workflow}}):

\begin{itemize}
    \item \textbf{Quota-based Search (Mode A):} When an explicit numerical target is detected (e.g., Find 5 papers about), the Counter operates as a sophisticated progress bar. It employs a semantic deduplication filter to verify entity uniqueness before incrementing. Crucially, it enforces an \textit{early-exit mechanism}: once the count reaches the target threshold $N$, the counter immediately terminates the agent's exploration loop, preventing redundant API costs.
    
    \item \textbf{Exhaustive Statistics (Mode B):} For open-ended inquiries (e.g., How many professors are listed?), the Counter switches to a coverage monitoring role. It drives the Explorer Agent to traverse the complete pagination chain until the environment is exhausted (i.e., no further ``Next Page'' signals), ensuring the final statistical output is grounded in full-scope observation rather than estimation.
\end{itemize}

By externalizing this logic, the framework ensures that the final result—whether a retrieved list or a statistical summary—is mathematically rigorous and robust against the stochastic nature of the LLM's internal memory.

\subsection{Hierarchical State Management via URL Stack}
\label{sec:url_stack}

Navigation in dynamic web environments is structurally equivalent to traversing a directed graph with unknown topology. A critical limitation in existing agents lack a cognitive map. They prone to getting stuck in loops or failing to backtrack correctly after exploring deep sub-pages, behaving like a browser without a ``Back'' button.

To address this, we engineer a dedicated \textbf{URL Stack} that serves as an external topological memory, enforcing a robust Depth-First Search (DFS) strategy. This module manages state transitions through two atomic operations:

\begin{itemize}
    \item \textbf{State Checkpointing (Push):} When the Explorer Agent decides to traverse a hyperlink to a child node, the system triggers a push operation. This snapshots the current url and strictly orders the exploration path (e.g., in \textbf{Round 1}), getting a reliable breadcrumb trail independent of the browser's history API.
    
    \item \textbf{Autonomous Error Recovery (Pop):} Upon rendering a new page, the Explorer Agent performs an immediate \textit{preliminary viability check}. If a page is detected—such as an HTTP 404 error or obviously irrelevant content then the explorer agent autonomously executes a pop operation. This mechanism instantly rolls back the environment to the previous valid state, effectively pruning invalid branches before the critic agent performs expensive semantic evaluation.
\end{itemize}

By formalizing navigation as stack-based state manipulation, our framework prevents cyclical loops and guarantees that the agent can systematically backtrack from local optima to explore alternative branches.

\subsection{Adaptive VLM Integration via US Calculator}

While VLMs excel at perceiving spatial layouts, their high computational overhead makes them impractical for every navigation step. To optimize the efficiency-accuracy trade-off, we introduce the US Calculator. This module dynamically evaluates the textual solvability of a page before the agent commits to a reasoning path.

As depicted in our framework (Figure \ref{fig:framework}), the policy $\pi$ adaptively switches modalities based on a learned threshold $\tau$:
\begin{equation}
    \pi(a_t | s_t) = 
    \begin{cases} 
    \pi_{\text{LLM}}(s_t) & \text{if } \Phi(s_t) > \tau \\
    \pi_{\text{VLM}}(s_t) & \text{otherwise}
    \end{cases}
\end{equation}
The composite score $\Phi(s_t)$ is the sum of four heuristic dimensions: Text Quality, Semantic Relevance, Structural Clarity and Special Case Penalties. Each designed to capture a specific failure mode of text-based DOM representations. A detailed analysis of these four dimensions is presented in Appendix~\ref{sec:us formulation}. By aggregating these scores, $\Phi(s_t)$ provides a rigorous justification for modality switching. If $\Phi(s_t) < \tau$, the \textbf{Critic Agent} recognizes that the text representation is a noise of the actual page and triggers the VLM pipeline to ensure grounded decision-making.

\section{Empirical Studies}
\label{sec:empirical studies}

\subsection{Constructing a Reproducible Benchmark: EverWebQA}
\begin{figure}[ht!]  
  \centering        
  \includegraphics[width=\columnwidth]{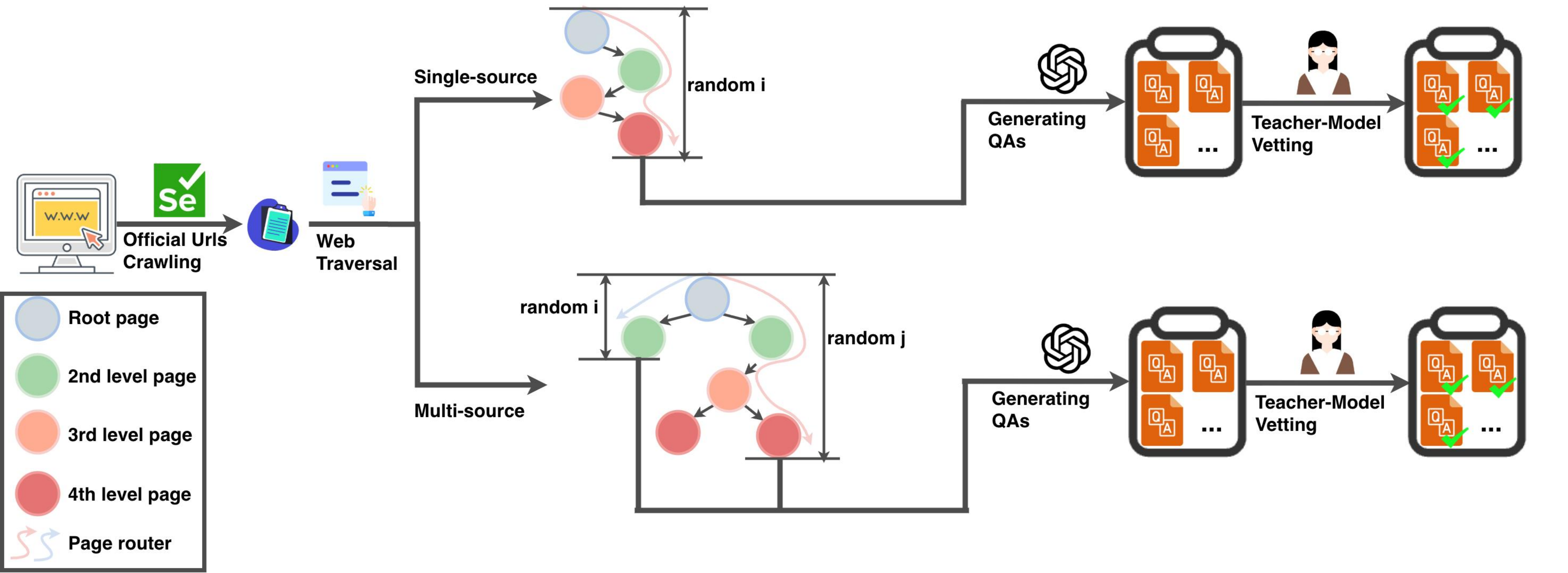} 
  \caption{The framework of test dataset generation}
  \label{fig:data_pipeline}
\end{figure}

To address the \textbf{temporal sensitivity} of web environments where static benchmarks rapidly suffer from content drift and link rot where we establish a reproducible data construction pipeline. Utilizing this pipeline, we introduce \text{EverWebQA}, a benchmark generated via a human-mimetic workflow that simulates navigational behaviors ranging from simple retrieval to complex synthesis. As illustrated in Figure \ref{fig:data_pipeline}, the framework integrates automated traversal with rigorous quality control to ensure sustainable and fair evaluation.

\paragraph{Hierarchical Web Traversal} 
We initiate the process by crawling 1,000 authoritative root URLs across four domains (Education, Conference, Organization, Game). Using Selenium, we reconstruct the DOM tree structure of each website, mapping nodes from the root page (Level 1) down to deep leaf nodes (Level 4), ensuring a robust structural foundation for traversal tasks.

\paragraph{Topology-Aware QA Synthesis} 
We simulate two distinct user navigation patterns to generate diverse queries:
\begin{itemize}[leftmargin=*]
    \item \textbf{Single-source}: Simulates a user drilling down to a specific topic. We sample a target node at depth $i \in [2, 4]$ (corresponding to Easy, Medium, Hard). The agent must navigate a linear path from the root to this specific depth to answer the query.
    \item \textbf{Multi-source}: Simulates a user integrating information across the site. We sample two disparate nodes at depths $i$ and $j$ within the \textbf{same} URL tree. The generated query necessitates reasoning over both pages, testing the agent's ability to maintain state across multi-branch navigation.
\end{itemize}

\paragraph{Teacher-Model Vetting} 
To eliminate hallucinations common in synthetic data, we employ a ``Teacher Model'' (GPT-4o) as a quality gatekeeper. As depicted in the final stage of Figure \ref{fig:data_pipeline}, the Teacher Model performs a rigorous \textit{solvability check}, verifying that the answer is explicitly derivable from the visited pages and refines the linguistic quality, retaining only high-fidelity samples for the final benchmark. The Teacher Model's prompt can see in~\ref{app:teacher model prompt}.

\paragraph{Data Statistics and Characteristics}
Following this structured data construction methodology, which combines $\text{LLM}$ guidance with programmatic web exploration, we constructed a final set of 680 question-answer pairs for $\text{EverWebQA}$. This specific volume was chosen to strictly align with the dataset size of the WebWalker baseline, thereby eliminating potential bias introduced by dataset scale discrepancies during comparison. Comprehensive statistics of $\text{EverWebQA}$, categorized by type, domain, and language, are provided below.

\begin{table}[htbp]
    \centering
    \small  
    \caption{Dataset Distribution by Question Type and Difficulty}
    \label{tab:dataset_distribution}
    \begin{tabular}{lccc}
        \toprule
        \textbf{Difficulty} & \textbf{Single-source} & \textbf{Multi-source} \\
        \midrule
        Easy   & 80 & 80 \\
        Medium & 140 & 140 \\
        Hard   & 120 & 120 \\
        \midrule
        \textbf{Total} & \textbf{340} & \textbf{340} \\
        \bottomrule
    \end{tabular}
\end{table}

\paragraph{Type and Difficulty Level}
\noindent Our dataset distinguishes between \textbf{Single-source} and \textbf{Multi-source} QAs. We define difficulty levels based on the navigational depth required to reach the target information. For \textbf{Single-source} tasks, a target page depth of 2, 3, and 4 corresponds to \textit{Easy}, \textit{Medium}, and \textit{Hard} levels, respectively. For \textbf{Multi-source} tasks, which necessitate synthesizing information from multiple pages, difficulty is determined by the \textit{cumulative depth} of all target pages: a total depth sum of 2--4 is classified as \textit{Easy}, 4--6 as \textit{Medium}, and 6--8 as \textit{Hard}.

\begin{figure}[ht!] 
    \centering
    \begin{subfigure}{0.48\linewidth}
        \centering
        \includegraphics[width=\linewidth]{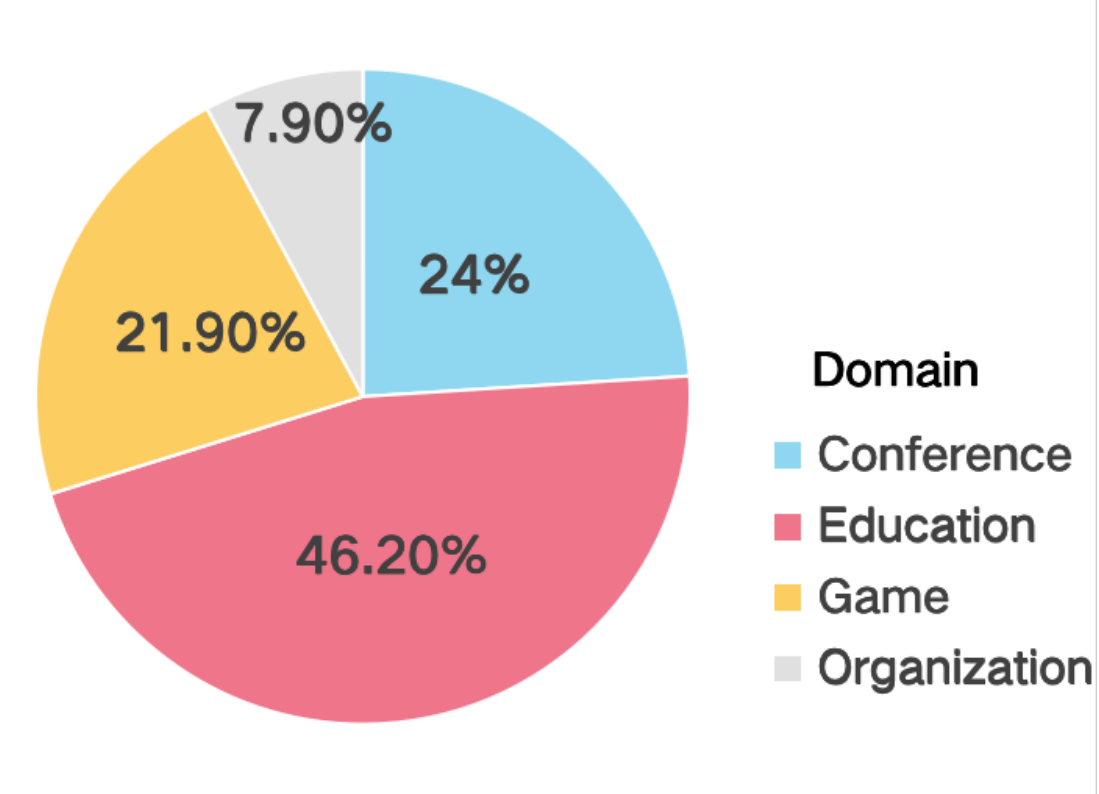}
        \caption{Domain distribution}
        \label{fig:domain_distribution}
    \end{subfigure}
    \hfill
    \begin{subfigure}{0.48\linewidth}
        \centering
        \includegraphics[width=\linewidth]{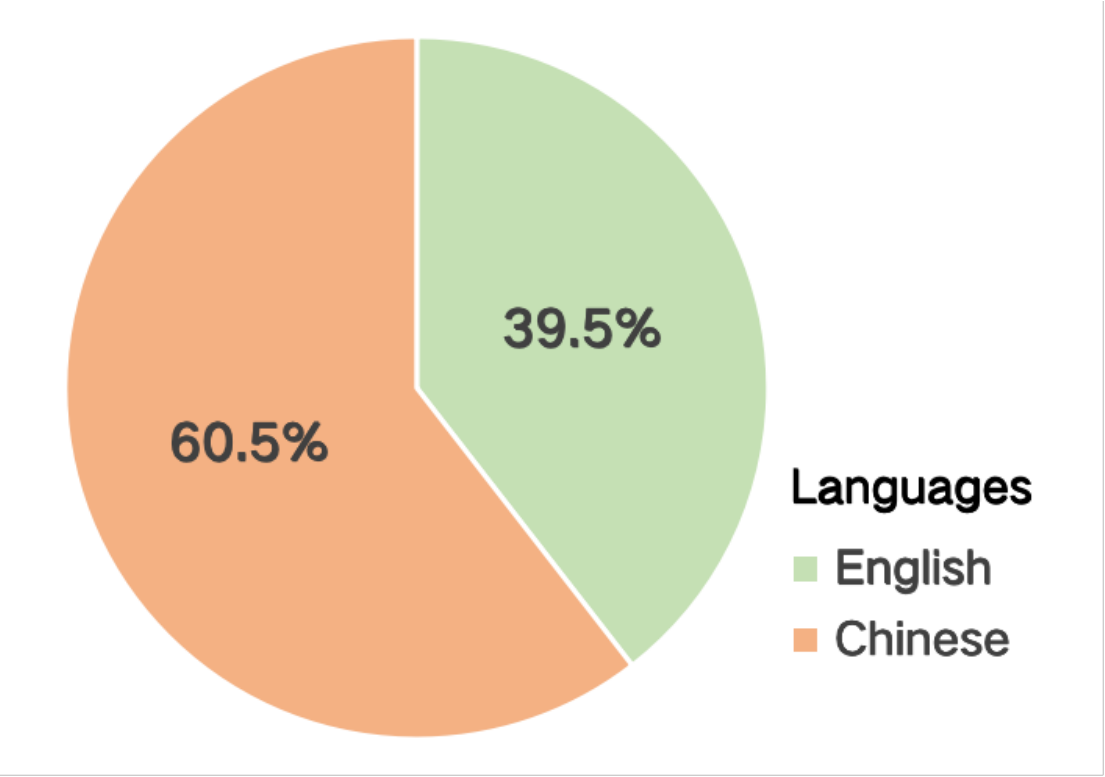}
        \caption{Language distribution}
        \label{fig:language_distribution}
    \end{subfigure}
    \caption{Data distribution of the EverWebQA dataset across domains and languages.}
    \label{fig:distribution}
\end{figure}

\paragraph{Domain and Language Distribution}
$\text{EverWebQA}$ is constructed across four real-world domains—education, conference, game, and organization—specifically selected for their authoritative content and complex structural depth. To reflect the linguistic diversity of the real-world web, the dataset is bilingual, featuring a substantial mix of both Chinese and English queries. As illustrated in Figure \ref{fig:distribution}, this diverse composition ensures a robust and comprehensive evaluation of $\text{V-GEM}$ across varying contexts and linguistic challenges.

\subsection{Baselines}
We adopt \textbf{WebWalker}~\cite{wu2025webwalker} as our baseline to ensure a rigorous and authoritative comparison. As a widely recognized state-of-the-art framework in autonomous web exploration, WebWalker represents the current standard for dual-agent architectures. However, it remains inherently constrained by its stochastic nature and lack of persistent memory.

By benchmarking against WebWalker, we aim to validate the specific contributions of the V-GEMS architecture. Rather than merely extending the baseline, V-GEMS is designed to address distinct limitations in arithmetic fidelity, navigational recall, and multi-modal reasoning. This comparison effectively isolates the impact of our architectural innovations, demonstrating that our modular enhancements yield significant performance improvements over the baseline system.

As in WebWalker, our system also operates in a zero-shot manner. To better evaluate practical applicability, we adopt Qwen-3-Coder-Plus as the underlying language model and adopt Qwen3-VL-Plus as the vision-language model. All experiments are conducted on the dataset generated on the day of experimentation~(November 25, 2025).\\

\subsection{Main Results}

\begin{figure}[ht!]
    \centering
    \subfloat[Comparison in type\label{fig:type_comparison}]{
        \includegraphics[width=0.9\linewidth]{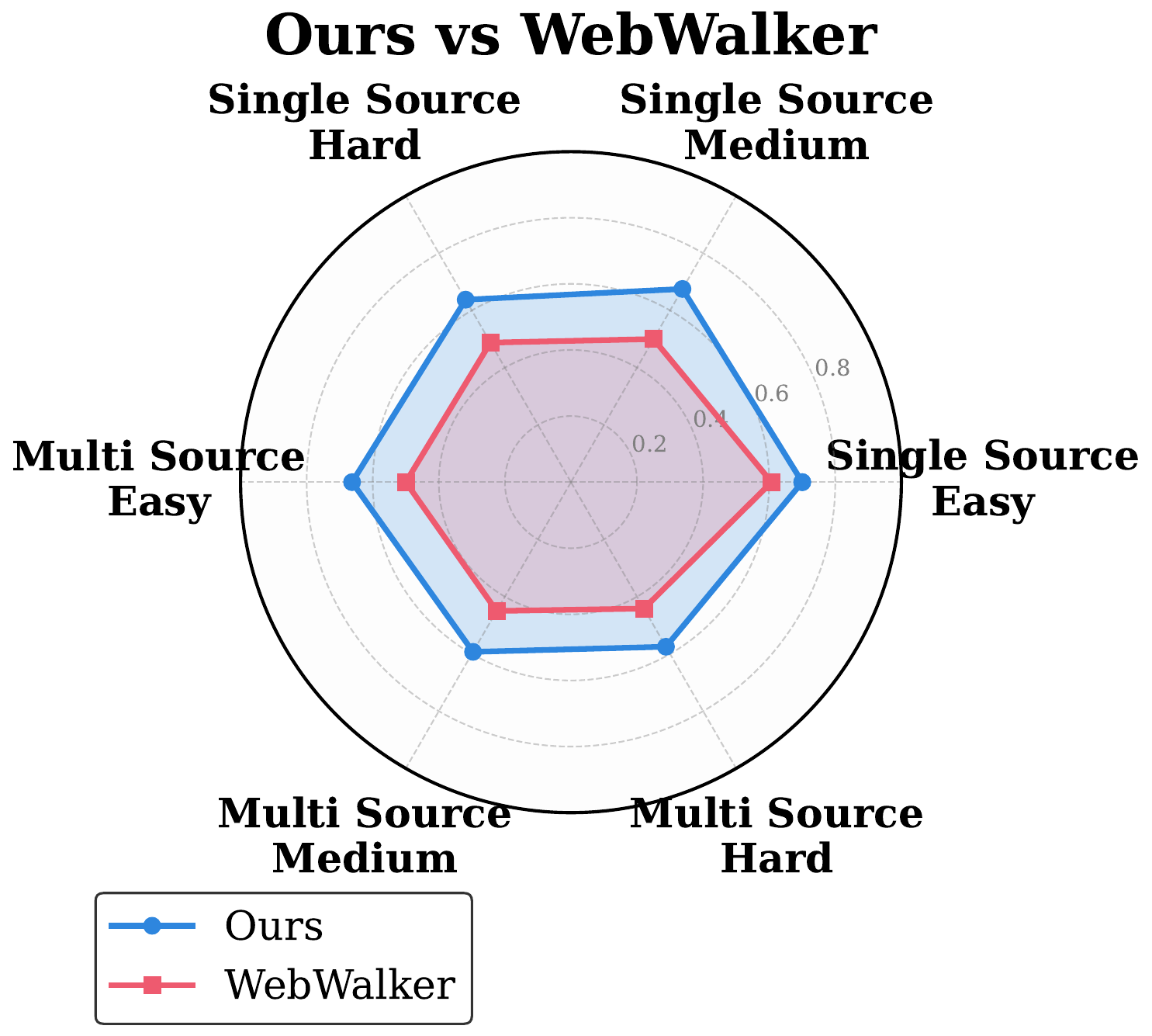}
    }
    \\ 
    \subfloat[Comparison in domain\label{fig:domain_comparison}]{
        \includegraphics[width=0.9\linewidth]{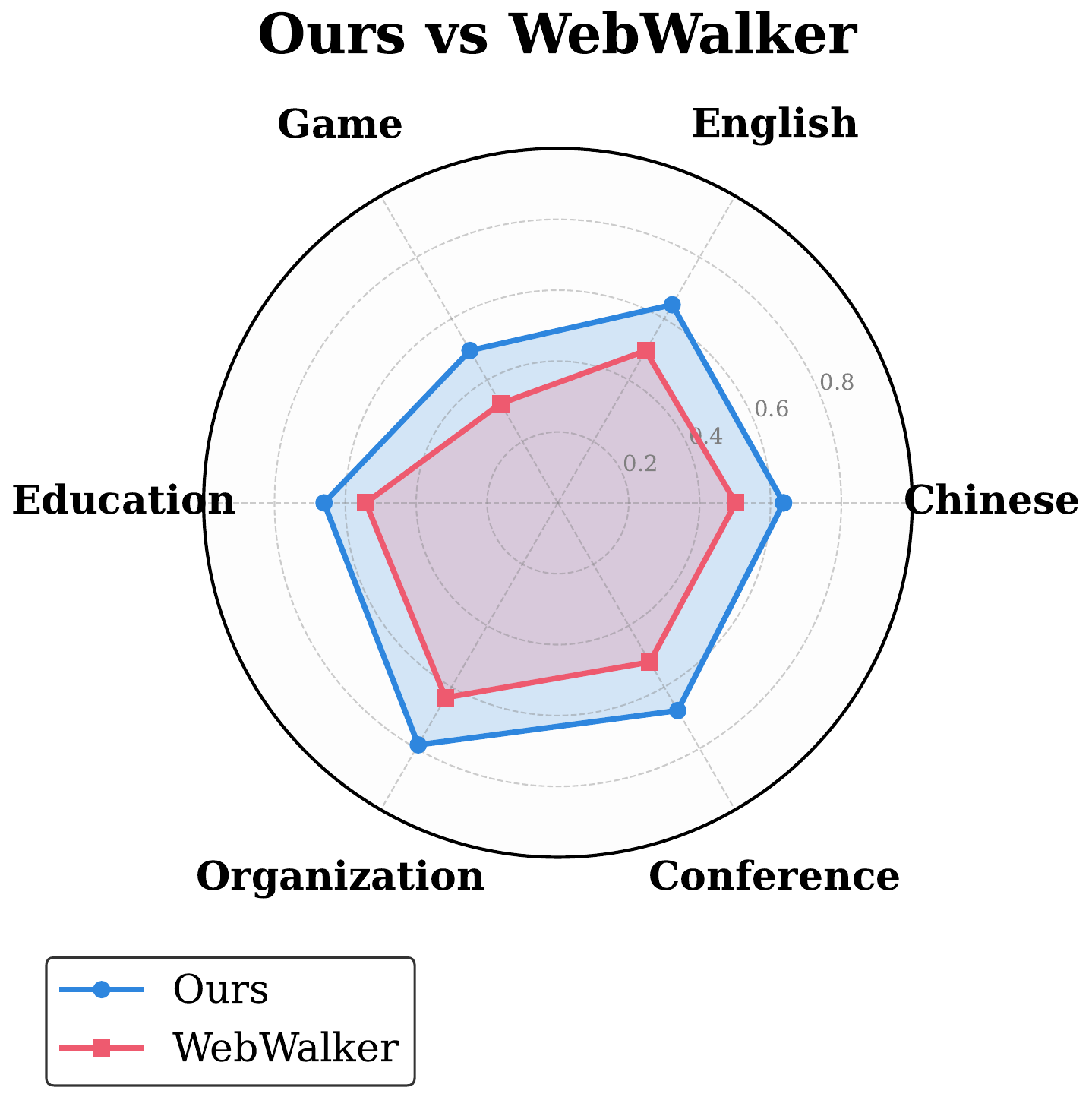}
    }
    
    \caption{The comparison in type and domain.}
    \label{fig:combined_comparison}
\end{figure}

Figure~\ref{fig:combined_comparison} presents a holistic performance evaluation. Overall, V-GEMS significantly outperforms the WebWalker baseline, elevating the average success rate from 0.49 to 0.65—a relative improvement of \textbf{28.7\%}.

\paragraph{Resilience to Complexity.}
Our model exhibits superior scaling laws regarding task difficulty. As shown in Figure~\ref{fig:type_comparison}, while the baseline struggles with \textit{Multi-Source Hard} tasks (due to memory loss during page transitions), V-GEMS maintains a high success rate. This substantial gap confirms that our \textbf{URL Stack} effectively provides a cognitive safety net, enabling the agent to synthesize information across disparate pages without losing navigational context. Similarly, the gains in \textit{Single-Source Medium/Hard} tasks validate the \textbf{Symbolic Counter}'s role in handling complex intra-page reasoning that typically confuses standard LLMs.

\paragraph{Domain Generalization.}
Figure~\ref{fig:domain_comparison} highlights the model's adaptability across structurally diverse environments. V-GEMS achieves dominant performance in the \textit{Conference}~(improved 33.33\%) and \textit{Game}~(improved 56.25\%) domains, which are characterized by irregular layouts and hybrid text-visual elements. This indicates that the \textbf{US Calculator} successfully dynamically switches to VLM perception when textual DOMs are insufficient, preventing the modal rigidity failures seen in WebWalker. Even in text-heavy domains like \textit{Education}, our agent maintains a clear lead, demonstrating that our architectural enhancements do not compromise basic text processing capabilities.

\subsection{Ablation Study}

\begin{table*}[ht!]
    \centering
    \small
    \renewcommand{\arraystretch}{1.25}
    \setlength{\tabcolsep}{4.5pt} 

    \caption{Ablation Test Results. The Modules columns indicate enabled components. The results are categorized by task granularity and domain type.}
    \label{tab:ablation_results_final}
    
    \begin{tabular}{ccc *{12}{c} c}
        \toprule
        \multicolumn{3}{c}{\textbf{Modules}} & 
        \multicolumn{6}{c}{\textbf{Fine-grained Tasks~(F1)}} & 
        \multicolumn{6}{c}{\textbf{Cross-domain Tasks~(F1)}} & 
        \multirow{3}{*}{\textbf{Avg}} \\ 
        
        \cmidrule(lr){1-3} \cmidrule(lr){4-9} \cmidrule(lr){10-15} 
        
        \multirow{2}{*}{\textbf{VLM}} & \multirow{2}{*}{\textbf{URL}} & \multirow{2}{*}{\textbf{Count}} & 
        
        \multicolumn{3}{c}{Single-Source} & \multicolumn{3}{c}{Multi-Source} & 
        
        \multirow{2}{*}{CN} & \multirow{2}{*}{EN} & \multirow{2}{*}{Game} & \multirow{2}{*}{Conf} & \multirow{2}{*}{Org} & \multirow{2}{*}{Edu} & 
        \\ 
        
        \cmidrule(lr){4-6} \cmidrule(lr){7-9}
        
         &  &  & 
        Easy & Med & Hard & Easy & Med & Hard & 
         &  &  &  &  &  & 
        \\
        \midrule
        
        - & - & - & 0.60 & 0.50 & 0.48 & 0.50 & 0.45 & 0.44 & 0.50 & 0.49 & 0.32 & 0.51 & 0.63 & 0.54 & 0.50 \\
        - & - & \checkmark & 0.63 & 0.52 & 0.48 & 0.49 & 0.46 & 0.46 & 0.51 & 0.50 & 0.34 & 0.51 & 0.63 & 0.55 & 0.51 \\
        - & \checkmark & - & 0.63 & 0.60 & 0.60 & 0.66 & 0.56 & 0.51 & 0.64 & 0.48 & 0.51 & 0.58 & 0.67 & 0.61 & 0.59 \\
        - & \checkmark & \checkmark & 0.65 & 0.61 & 0.59 & 0.67 & 0.57 & 0.52 & 0.63 & 0.50 & 0.52 & 0.59 & 0.69 & 0.61 & 0.59 \\
        \checkmark & - & - & 0.67 & 0.63 & 0.61 & 0.64 & 0.59 & 0.55 & 0.62 & 0.61 & 0.47 & 0.66 & 0.73 & 0.63 & 0.62 \\
        \checkmark & - & \checkmark & 0.68 & 0.66 & 0.61 & 0.65 & 0.55 & 0.53 & 0.62 & 0.60 & 0.46 & 0.64 & 0.75 & 0.64 & 0.62 \\
        \checkmark & \checkmark & - & 0.69 & 0.67 & 0.63 & 0.63 & 0.60 & 0.57 & 0.63 & 0.64 & 0.50 & 0.68 & 0.79 & 0.64 & 0.64 \\
        \checkmark & \checkmark & \checkmark & \textbf{0.70} & \textbf{0.68} & \textbf{0.64} & \textbf{0.66} & \textbf{0.60} & \textbf{0.57} & \textbf{0.64} & \textbf{0.64} & \textbf{0.50} & \textbf{0.68} & \textbf{0.79} & \textbf{0.66} & \textbf{0.65} \\
        \bottomrule
    \end{tabular}
\end{table*}

To investigate the individual contribution of each proposed module, we conduct extensive ablation tests across fine-grained and cross-domain tasks. Table~\ref{tab:ablation_results_final} summarizes the performance for various configurations.

\paragraph{Quantifying Contributions via Shapley Value.}
To provide a rigorous interpretation of component utility, we employ the \textit{Shapley Value}~\citep{shapley1953value} metric to quantify the marginal contribution of each module. Formally, we define our three-module system as $\mathcal{M} = \{C, S, U\}$, where $C$ denotes the \textit{Symbolic Counter}, $S$ represents the \textit{URL Stack}, and $U$ stands for the \textit{Understanding Calculator}.
Taking the Symbolic Counter ($C$) as an example, its contribution $\phi_C$ for a specific task is calculated as the weighted average of its marginal gains across all subset combinations:
\begin{equation}
    \phi_C = \frac{1}{3} \left[ \Delta_{C} + \frac{\Delta_{C|S} + \Delta_{C|U}}{2} + \Delta_{C|SU} \right] \times 10^3
\end{equation}
where $\Delta_{C} = (Acc_{\{C\}} - Acc_{\emptyset})$ represents the primary gain, $\Delta_{C|S}$ and $\Delta_{C|U}$ denote the marginal gains when added to single-module baselines (e.g., $Acc_{\{C,S\}} - Acc_{\{S\}}$), and $\Delta_{C|SU} = (Acc_{\{C,S,U\}} - Acc_{\{S,U\}})$ captures the gain within the full system. We scale the final values by $10^3$ to enhance visual clarity in our analysis.

\begin{figure}[ht!] 
    \centering
    \subfloat[Shapley Value in type]{\label{fig:task_type_shapley_grouped}
        \includegraphics[width=0.9\linewidth]{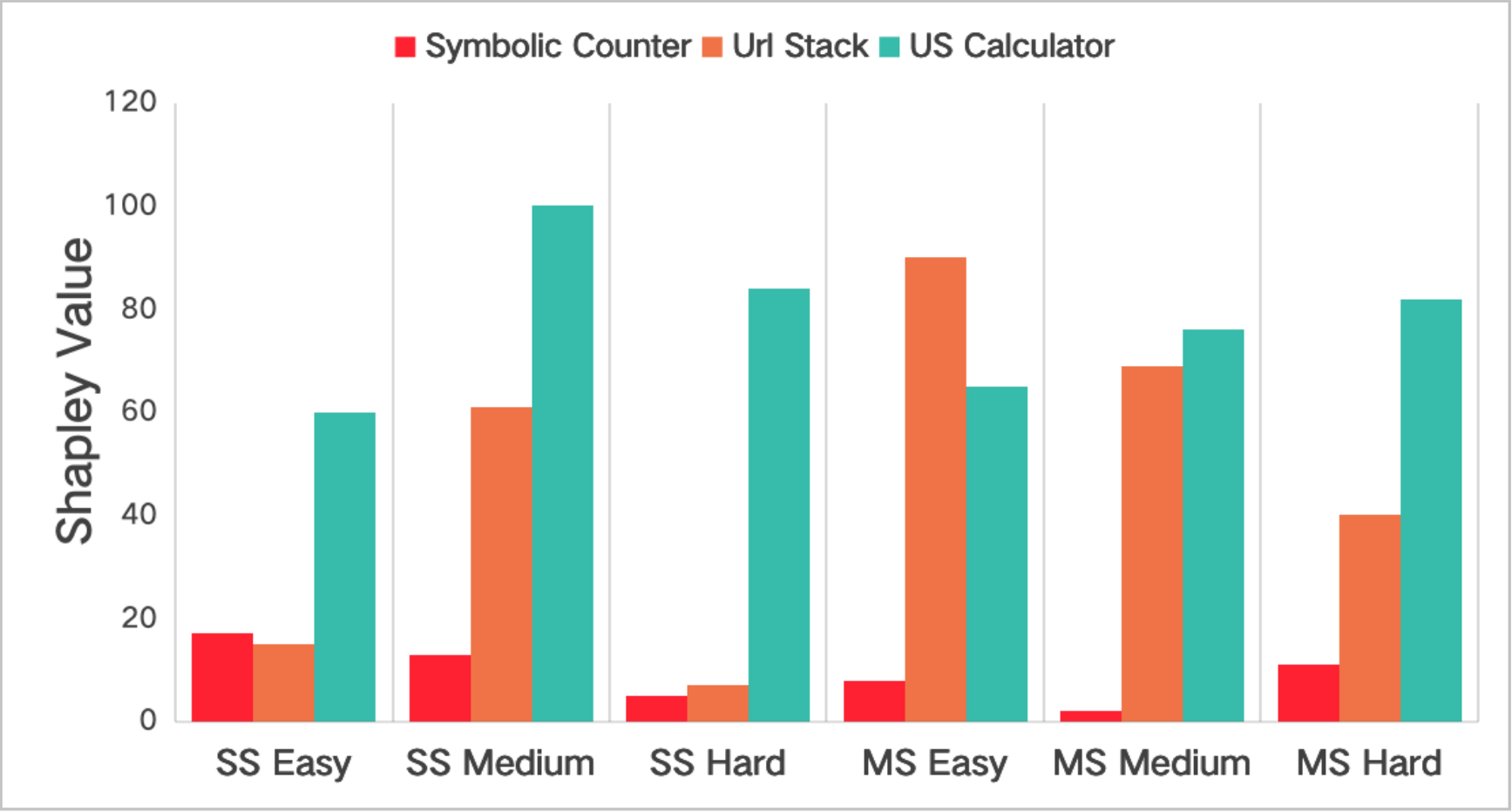}
    }
    \\
    \subfloat[Shapley Value in domain]{\label{fig:domain_shapley_grouped}
        \includegraphics[width=0.9\linewidth]{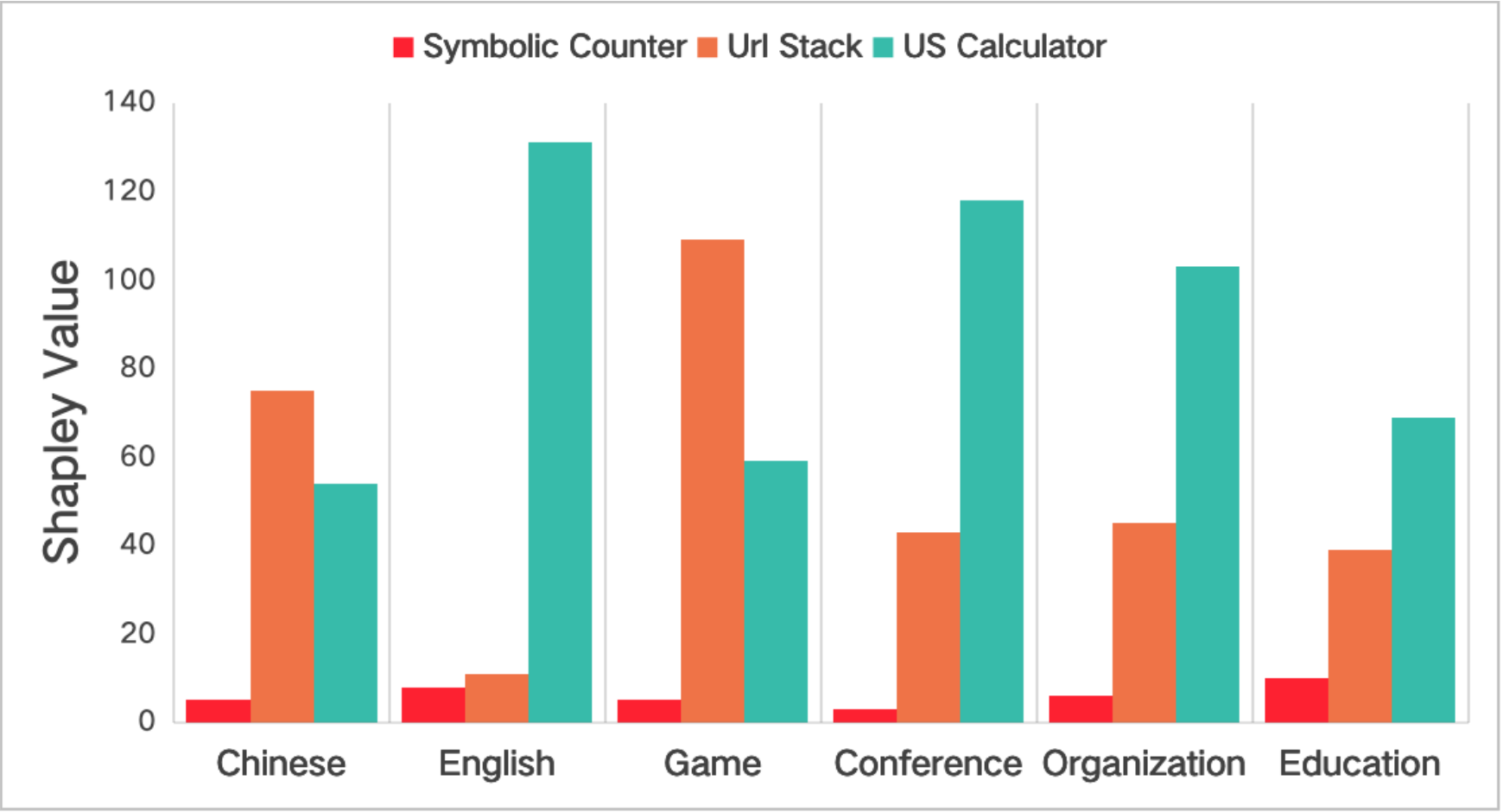}
    }
    \caption{Three methods' contribution in different task}
    \label{fig:combined_shapley_grouped}
\end{figure}

\paragraph{Task-wise Insight: Perception vs. Navigation.}
The Shapley values across task types (see Figure~\ref{fig:task_type_shapley_grouped}) reveal distinct functional specializations:
\begin{itemize}
    \item \textbf{VLM Dominance in SS Tasks}: The VLM module exhibits the highest contribution in all \textit{Single-Source (SS)} tasks, peaking at nearly 100 in \textit{SS Medium}. This underscores that visual perception is the primary bottleneck when navigating individual pages with complex layouts.
    \item \textbf{URL Stack in MS Exploration}: For \textit{Multi-Source (MS)} tasks, the contribution of the URL Stack increases significantly, reaching its zenith (90) in \textit{MS Easy} scenarios. This validates our hypothesis that stateful backtracking is indispensable for tasks requiring cross-page information retrieval.
\end{itemize}

\paragraph{Domain-wise Insight: Structural vs. Semantic Complexity.}
Analysis of domain-specific Shapley values (Figure~\ref{fig:domain_shapley_grouped}) further highlights the adaptive nature of our framework:
\begin{itemize}
    \item \textbf{Visual Primacy}: In the English and Conference domains, the VLM module shows extraordinary Shapley values (130 and 118 respectively). These domains typically feature information-dense visual banners and complex navigation bars that textual DOMs fail to represent accurately.
    \item \textbf{Structural Navigation}: Conversely, in the Game and Chinese domains, the URL Stack emerges as a leading contributor (110 and 78 respectively). This suggests that these domains rely more on traversing deep hierarchical structures where the ability to ``pop'' back to previous states is more critical than raw visual perception.
    \item \textbf{Counter Stability}: Although the Counter maintains a lower overall Shapley value, it provides a stable contribution (10) in the Education domain, reflecting its role in ensuring arithmetic precision during specific data-aggregation tasks.
\end{itemize}


\subsection{Error Assessment}
\begin{figure}[ht] 
    \centering
    \includegraphics[width=\linewidth]{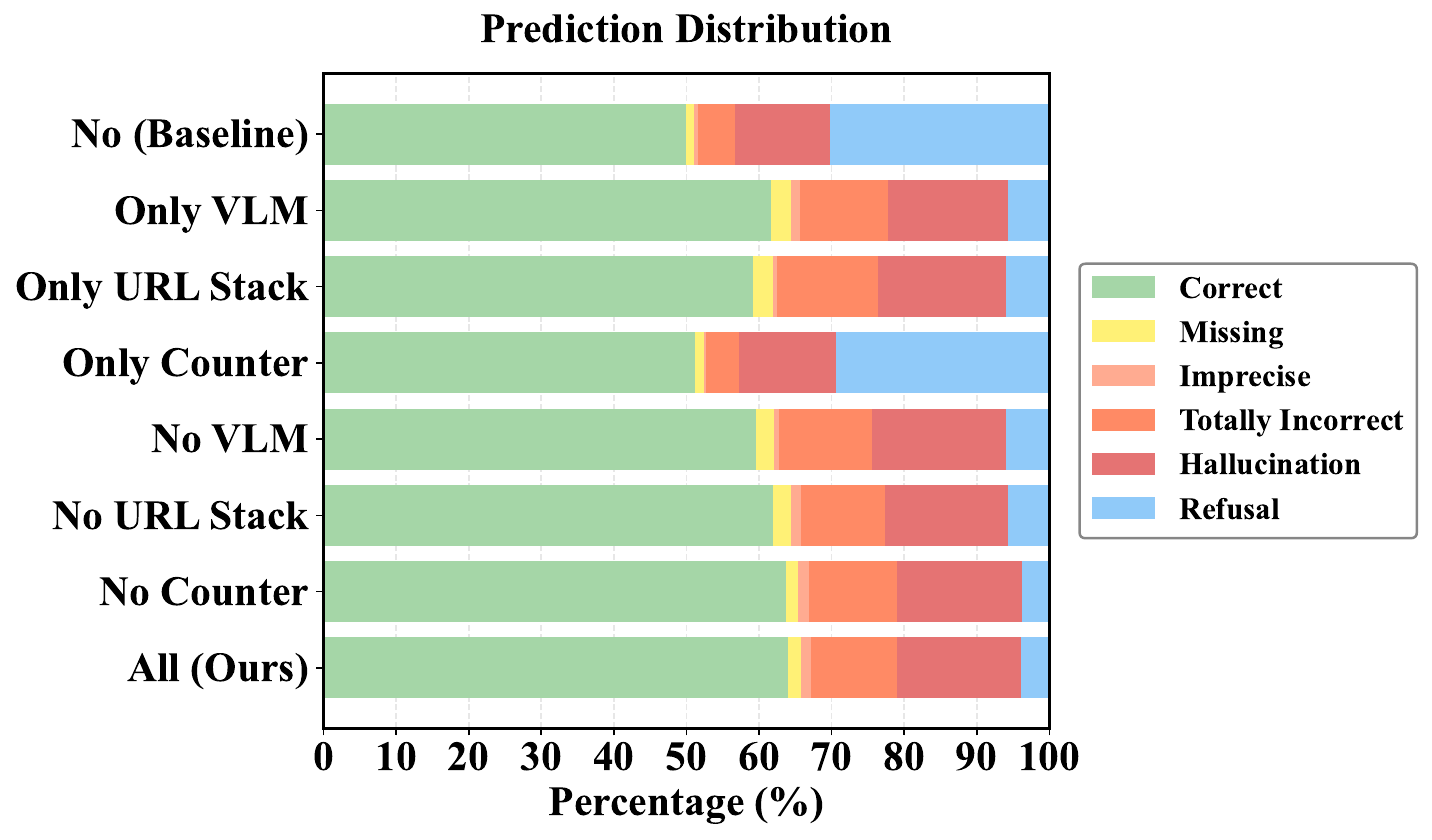} 
    \caption{Fine-grained error distribution across model variants. Our full system (\textbf{All}) significantly improves the correctness rate while suppressing severe failures like Refusal and Hallucination.}
    \label{fig:error_distribution} 
\end{figure}

As visualized in the prediction distribution (Figure~\ref{fig:error_distribution}), we categorize failures into five distinct modes to dissect performance bottlenecks. The baseline variants exhibit a pronounced tendency toward Refusal and Totally Incorrect outcomes, characterizing a form of navigational problem where agents prematurely terminate exploration upon hitting dead ends. Introducing the URL Stack effectively suppresses this behavior, providing a cognitive safety net that converts navigational failures into successful retrievals. Furthermore, while stochastic hallucination remains an inherent challenge for LLMs, the integration of the US Calculator and Symbolic Counter significantly thins the Hallucination segment by decoupling perceptual and arithmetic reasoning from the generative process. Despite these architectural gains, the persistence of Missing Key Info errors even when correct pages are visited underscores a residual \textit{retrieval-to-synthesis gap}, where the agent locates target information but struggles to exhaustively aggregate fine-grained details under context constraints.

\section{Conclusion}
\label{sec:conclusion}

In this work, we presented \textbf{V-GEMS}, a multimodal web agent designed to resolve the navigational amnesia and arithmetic hallucinations inherent in LLM-based exploration. By augmenting the framework with a Symbolic Counter, URL Stack, and an adaptive US Calculator, and benchmarking on our reproducible EverWebQA dataset, we achieved a significant 28.7\% performance gain over the WebWalker baseline. These results confirm that decoupling specific cognitive functions from stochastic generation is crucial for building reliable, efficient web agents.

Future work will prioritize efficiency and open-ended autonomy. We plan to distill our architecture into lightweight, specialized models to reduce computational latency and integrate Agentic Reinforcement Learning to optimize long-horizon trajectories. Ultimately, we aim to eliminate the dependency on pre-defined root URLs, empowering the agent to initiate open-world information seeking from generalized starting points.

\section*{Limitations}

While V-GEMS demonstrates robust performance in multimodal web navigation, we identify three primary limitations in our current implementation:

\paragraph{Computational Overhead and Latency.}
The integration of VLMs via the US Calculator introduces a trade-off between perception accuracy and inference speed. Although our adaptive switching mechanism reduces unnecessary calls, the multi-modal processing pipeline still incurs higher latency and token costs compared to purely text-based baselines. This currently limits the system's applicability in time-sensitive, low-resource scenarios. We plan to address this via model distillation and quantization to compress the agent into a more lightweight form without sacrificing performance.

\paragraph{Dependency on Pre-defined Entry Points.}
Our current evaluation protocol relies on a specific root url to initiate exploration. This setup does not fully replicate open-ended internet surfing, where an agent must start from a general search engine without a known domain boundary. Generalizing V-GEMS to initiate and plan tasks from an unbounded starting point remains an open challenge for achieving true open-world autonomy.

\paragraph{Absence of Self-Evolving Optimization.}
The decision-making logic of V-GEMS currently relies on heuristic-based planning and in-context learning. It lacks an end-to-end Reinforcement Learning (RL) mechanism to update its policy based on trial-and-error feedback. Consequently, the agent may not always converge on the globally optimal navigation trajectory in highly complex environments. Integrating Agentic RL to enable self-improving capabilities is a critical direction for our future research.



\bibliography{custom, biblio}

\appendix

\section{Detailed Formulation of VLM Understanding Score}
\label{sec:us formulation}
\subsection{Text Quality ($S_{\text{qual}}$)} 
This metric assesses whether the DOM contains enough legible information to be interpreted by an LLM. It is defined as:
\begin{equation}
    S_{\text{qual}} = f_{\text{len}}(|O|) + 10 \cdot \frac{|O|_{\text{valid}}}{|O|} + S_{\text{fmt}}
\end{equation}
where $|O|$ is the total character count, and $|O|_{\text{valid}}$ represents the count of non-gibberish, meaningful characters. The function $f_{\text{len}}$ acts as a "Goldilocks" filter: it penalizes sparse pages (likely containing only images) and excessively bloated pages (likely containing junk scripts). $S_{\text{fmt}}$ rewards the presence of structural HTML tags (like \texttt{<li>} or \texttt{<h1>}), which are crucial for LLM understanding.

\subsection{Semantic Relevance ($S_{\text{rel}}$)} 
This component (max 40 pts) ensures the current page actually contains content related to the user query $q$. We utilize a lightweight keyword-overlap and semantic embedding check between $q$ and the textual observation $O$. If $S_{\text{rel}}$ is low, it indicates that the textual content is irrelevant, prompting the agent to invoke the VLM to see if the target information is embedded in visual banners or icons instead. Detailed prompt can see in~\ref{app:rel prompt}

\subsection{Structural Clarity ($S_{\text{struct}}$)} 
Navigability is often hindered by cluttered or empty DOM trees. We quantify this as:
\begin{equation}
    S_{\text{struct}} = 5 \cdot \mathbb{I}(n_{\text{para}} \geq 3) + f_{\text{nav}}(n_{\text{btn}}) + S_{\text{dense}}
\end{equation}
where $n_{\text{para}}$ is the number of text paragraphs and $\mathbb{I}(\cdot)$ is the indicator function. $n_{\text{btn}}$ denotes the count of interactive elements (buttons/links). The function $f_{\text{nav}}$ peaks when $n_{\text{btn}} \in [5, 30]$; too few buttons suggest a dead-end page, while too many suggest a complex navigation bar that requires visual spatial reasoning ($S_{\text{dense}}$) to disambiguate.

\subsection{Special Case Penalties ($S_{\text{spec}}$)} 
To handle scenarios where text is inherently insufficient, we apply a baseline score and subtract penalties $p_i$ for visual-only elements:
\begin{equation}
    S_{\text{spec}} = \max\left(0, 15 - \sum_{i \in \mathcal{K}} p_i\right)
\end{equation}
The set $\mathcal{K}$ includes keywords detected in the DOM metadata indicating image galleries ($p_1$), captchas ($p_2$), or complex error pages ($p_3$).

\section{Comprehensive Workflow Demonstration}
\label{sec:demonstration}
\begin{figure*}[t!]
    \centering
    \includegraphics[width=0.95\textwidth]{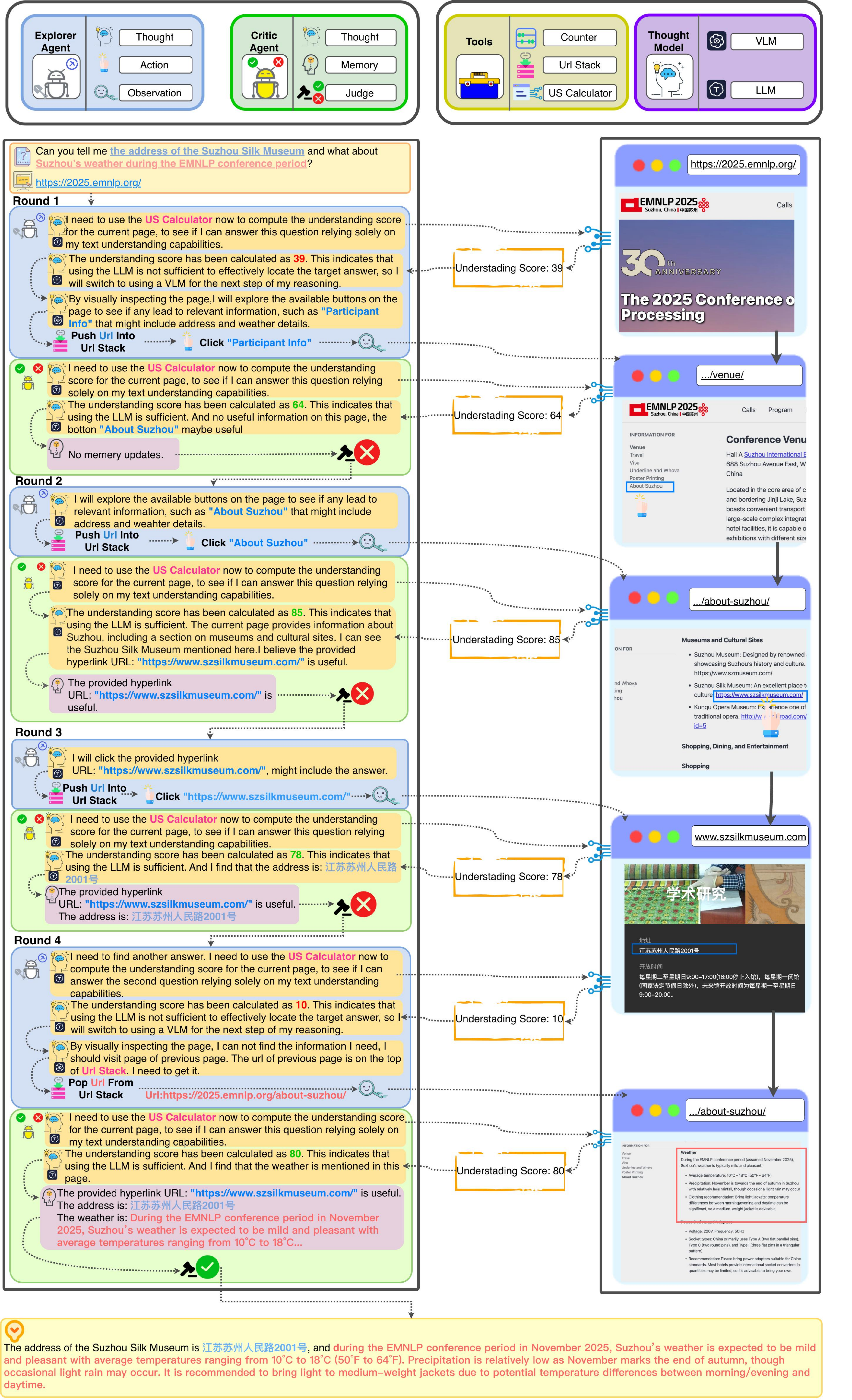} 
    \vspace{-0.3cm}
    \caption{A comprehensive workflow demonstration.}
    \label{fig:poster}
\end{figure*}

To provide a concrete visualization of the V-GEMS inference pipeline, Figure \ref{fig:poster} presents a step-by-step execution trace on a complex multi-source query. This qualitative example decomposes the agent's trajectory into distinct phases—ranging from initial query analysis to recursive exploration and final answer synthesis.

\section{Detailed Workflow of the Symbolic Counter}
\label{sec:appendix_counter_workflow}

\begin{figure}[t!] 
    \centering
    \vspace*{-7cm}
    \includegraphics[width=0.95\columnwidth]{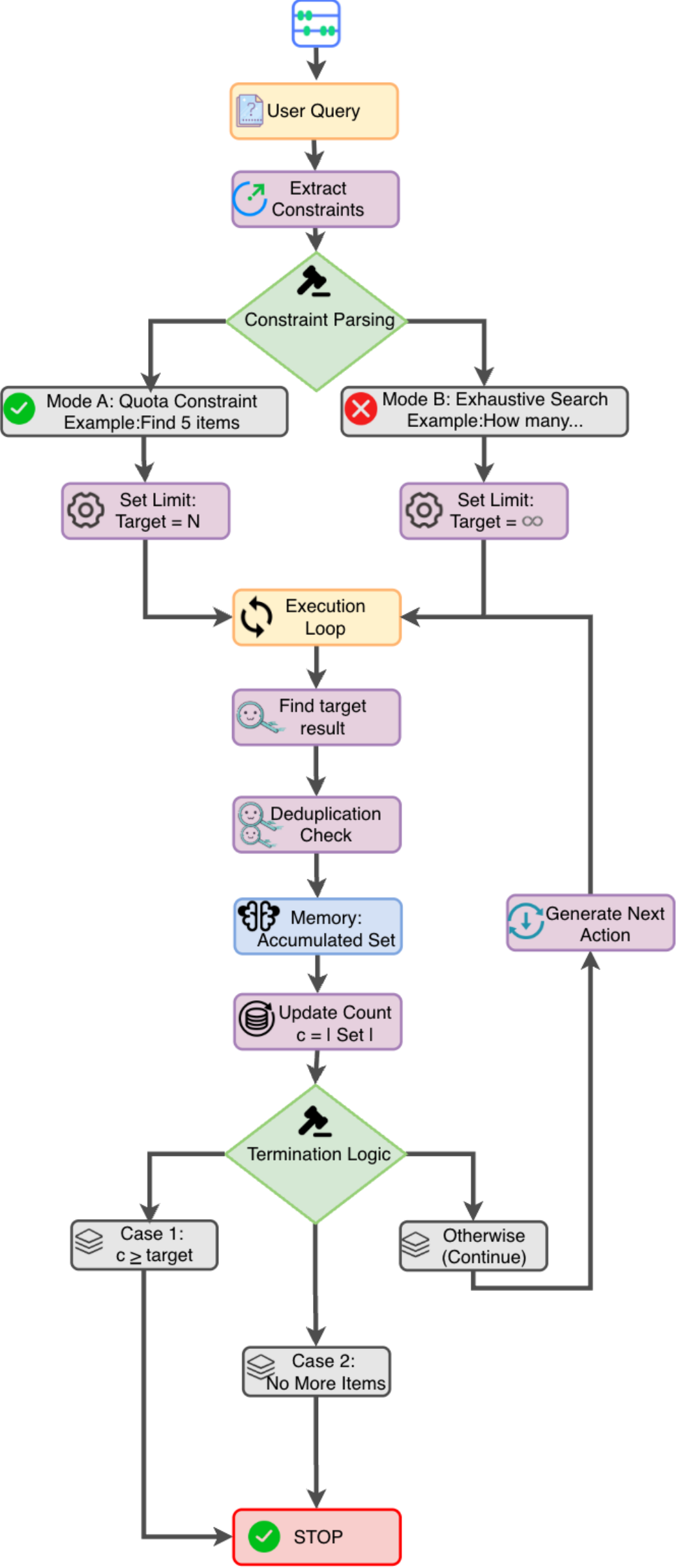} 
    \caption{Operational flowchart of the Symbolic Counter.}
    \label{fig:appendix_counter} 
\end{figure}

Complementing the methodology described in Section \ref{sec:counter}, Figure \ref{fig:appendix_counter} presents the comprehensive algorithmic flowchart of the \textbf{Symbolic Counter}. The diagram visually details the branching logic for constraint parsing, entity deduplication, and the specific termination conditions for both quota-based and exhaustive search modes.

\section{Details on Prompts}
\label{sec:prompts}

\subsection{Prompts for Evaluation Metrics}
To ensure the reproducibility of our experimental results, we detail the specific system instructions employed in our LLM-as-a-Judge evaluation framework. The following prompts quantify the agent's performance across three dimensions: computing the $S_{\text{rel}}$ for understanding score, verifying correctness to determine the overall success rate, and performing error classification for failure analysis.

\begin{tcolorbox}[
enhanced jigsaw,
breakable,
pad at break*=1mm,
colback=white!95!gray,
colframe=gray!50!black,
title={Prompt for $S_{\text{rel}}$ Calculation}]
\label{app:rel prompt}
\begin{lstlisting}[style=promptstyle]
You are an expert in information retrieval relevance assessment.
Evaluate the relevance between the User Query and the Webpage Content.

Input:
- User Query: (*@\textcolor{blue}{\{query\}}@*)
- Webpage Content: (*@\textcolor{blue}{\{observation\}}@*)

Scoring Criteria (0-40):
- 40: Highly Relevant (Direct answer or key evidence).
- 30: Relevant (Useful supporting info or clear cues).
- 20: Weakly Relevant (Indirect links, tangential).
- 10: Marginally Relevant (Mostly noise).
- 0: Irrelevant.

Output JSON:
{
    "score": <integer 0-40>,
    "reason": "<brief justification>"
}
\end{lstlisting}
\end{tcolorbox}

\begin{tcolorbox}[
enhanced jigsaw,
breakable,
pad at break*=1mm,
colback=white!95!gray,
colframe=gray!50!black,
title={Prompt for Evaluation}]
\begin{lstlisting}[style=promptstyle]
You are an expert evaluator for question-answering systems.
Given a question, a reference answer, and a prediction, your task is to determine if the prediction correctly answers the question based on the reference answer.

Input Data:
- Question: (*@\textcolor{blue}{\{question\}}@*)
- Reference Answer: (*@\textcolor{blue}{\{answer\}}@*)
- Prediction: (*@\textcolor{blue}{\{pred\}}@*)

Evaluation Criteria:
1. Compare the prediction with the reference answer strictly.
2. The prediction must contain the key entities/facts from the reference.
3. Minor wording differences are acceptable if the semantic meaning is identical.
4. If the prediction is wrong, incomplete (missing key parts), or fails to answer, score it as INCORRECT (0).

Output Format:
Please provide your evaluation in the following format:
REASONING: [Detailed reasoning about why it is correct/incorrect]
SCORE: [1 or 0]

Your evaluation:
\end{lstlisting}
\end{tcolorbox}

\begin{tcolorbox}[
enhanced jigsaw,
breakable,
pad at break*=1mm,
colback=white!95!gray,
colframe=gray!50!black,
title={Prompt for Error Classification}]
\begin{lstlisting}[style=promptstyle]
You are a rigorous QA evaluator.
Your task is to classify the error in the Model Prediction given the Question, Reference Answer, and the retrieved Web Information.

Input Data:
- Question: (*@\textcolor{blue}{\{question\}}@*)
- Reference Answer (Ground Truth): (*@\textcolor{blue}{\{answer\}}@*)
- Model Prediction: (*@\textcolor{blue}{\{pred\}}@*)
- Web Information (Context): (*@\textcolor{blue}{\{information\}}@*)

Error Taxonomy (Choose strictly one):

1. [Hallucination] (Fabrication)
   - Definition: The prediction asserts specific facts (dates, numbers, names) that appear NOWHERE in the provided "Web Information".
   - Key Check: Is this information completely made up or extrinsic to the source?

2. [Totally Incorrect] (Reasoning Error)
   - Definition: The prediction is factually wrong, but the error stems from misinterpreting the "Web Information". The entity exists in the text, but the model selected the wrong one (e.g., confusing "Submission Deadline" with "Notification Date").
   - Key Check: Is it a logic/mapping error rather than pure fabrication?

3. [Missing Key Info] (Incompleteness)
   - Definition: The prediction is factually correct but incomplete. It fails to list all items required (Recall failure) or omits a specific component of a multi-part query (e.g., answering the "who" but missing the "where").

4. [Imprecise] (Granularity Mismatch)
   - Definition: The prediction is correct in direction but lacks the required specific resolution.
   - Example: Answering "July 2023" when the truth is "July 21, 2023". It is vague rather than wrong.

Instruction:
Compare the Prediction against the Reference and Context.
Output ONLY the category name.

Output:(Choose one: Hallucination or Totally Incorrect or Missing Key Info or Imprecise. Don't output anything else.)
\end{lstlisting}
\end{tcolorbox}

\subsection{Prompts for Agent Modules}
This section details the system instructions for the core components of the V-GEMS framework. 

\begin{tcolorbox}[
enhanced jigsaw,
breakable,
pad at break*=1mm,
colback=white!95!gray,
colframe=gray!50!black,
title={Prompt for Explorer Agent}]
\begin{lstlisting}[style=promptstyle]
You are a web exploration agent. Your goal is to find quality information by navigating through websites to answer the user's query.

Available Tools:
(*@\textcolor{blue}{\{tool\_descs\}}@*)

(*@\textbf{[Operational Protocols]}@*)
1. Counter Enforcement (CRITICAL):
   - You must collect a specific number of items.
   - ALWAYS check progress using (*@\texttt{count\_usefulness}@*).
   - Rule: IF (*@\texttt{current\_count}@*) < (*@\texttt{target\_count}@*): You MUST CONTINUE exploring. DO NOT STOP.
   - Rule: IF (*@\texttt{current\_count}@*) >= (*@\texttt{target\_count}@*): You may proceed to generate the final answer.

2. State Management:
   - Loop Prevention: Before acting, check the (*@\texttt{url\_stack}@*). NEVER click a link that leads to a visited URL.
   - Backtracking: If the current branch is exhausted or irrelevant, use (*@\texttt{url\_stack}@*) to backtrack or jump to a sibling node.

(*@\textbf{[Format Requirement]}@*)
Use the following format strictly:

Question: (*@\textcolor{blue}{\{query\}}@*)
Thought: Analyze the current state.
Action: The action to take, must be one of [(*@\textcolor{blue}{\{tool\_names\}}@*)].
Action Input: The input to the action (JSON format).
Observation: The result of the action.
... (Repeat Thought/Action/Observation loop until success)

Begin!
\end{lstlisting}
\end{tcolorbox}

\begin{tcolorbox}[
enhanced jigsaw,
breakable,
pad at break*=1mm,
colback=white!95!gray,
colframe=gray!50!black,
title={Prompt for Critic Agent}]
\begin{lstlisting}[style=promptstyle]
You are a Critic Agent responsible for information assessment and task termination. This process operates in two sequential stages.

(*@\textbf{[Stage 1: Information Filtering]}@*)
Task: Analyze the current observation. If it contains information relevant to the query, extract it precisely. Eliminate noise.

Input:
- Query: (*@\textcolor{blue}{\{query\}}@*)
- Current Observation: (*@\textcolor{blue}{\{observation\}}@*)

Output JSON:
{
    "is_useful": <true/false>,
    "extracted_info": "<Specific details extracted from observation, or null>"
}

(*@\textbf{[Stage 2: Sufficiency Judgment]}@*)
Task: Evaluate the global accumulated information. Decide if it is sufficient to fully answer the user's query.

Input:
- Query: (*@\textcolor{blue}{\{query\}}@*)
- Accumulated Info: (*@\textcolor{blue}{\{accumulated\_info\}}@*) (Aggregated from Stage 1 history)

Output JSON:
{
    "is_sufficient": <true/false>,
    "final_answer": "<Generate the final answer if sufficient, otherwise null>"
}
\end{lstlisting}
\end{tcolorbox}

\subsection{Prompts for Data Synthesis}
Finally, we detail the system instructions used for our automated data synthesis pipeline. 

\begin{tcolorbox}[
enhanced jigsaw,
breakable,
pad at break*=1mm,
colback=white!95!gray,
colframe=gray!50!black,
title={Prompt for QA Pair Generation}]

\begin{lstlisting}[style=promptstyle]
(*@\textbf{[Mode A: Single-Source QA Generation]}@*)
You are a professional QA dataset generation assistant. I will give you a web page's content, please generate a QA pair based on the information from this page.

Website domain: (*@\textcolor{blue}{\{domain\}}@*)
(*@\textcolor{blue}{\{domain\_instruction\}}@*)

(*@\textbf{Quality Requirements (VERY IMPORTANT):}@*)
1. Questions MUST have clear, verifiable, objective answers (e.g., specific dates, locations, names, numbers, concrete facts)
2. Avoid subjective questions (e.g., "How good is...", "What do you think about...", "How would you rate...")
3. Answers must come directly from the page content, do not make up or speculate
4. Question type examples:
   (*@$\checkmark$@*) Good questions: "When was the university founded?" "What is the paper submission deadline?" "How many players does the game support?"
   (*@$\times$@*) Avoid: "How is the university?" "What are the conference features?" "Is the game fun?"
5. Answers should be concise and accurate with specific information, not vague or general descriptions

(*@\textbf{Requirements:}@*)
1. The question should be answerable from this page's content
2. The question should be related to the (*@\textcolor{blue}{\{domain\}}@*) domain
3. Difficulty level: (*@\textcolor{blue}{\{difficulty\}}@*)
   - easy: Information is direct and obvious, easy to find
   - medium: Requires understanding and integration of information
   - hard: Requires in-depth analysis and reasoning
4. Generate question and answer in English

Input Content:
(*@\textcolor{blue}{\{content\}}@*)

(*@\centerline{\rule{0.8\linewidth}{0.5pt}}@*) 

(*@\textbf{[Mode B: Multi-Source QA Generation]}@*)
You are a professional QA dataset generation assistant. I will give you content from two web pages, please generate a QA pair based on the information from BOTH pages.

Website domain: (*@\textcolor{blue}{\{domain\}}@*)
(*@\textcolor{blue}{\{domain\_instruction\}}@*)

(*@\textbf{Quality Requirements (VERY IMPORTANT):}@*)
1. Questions MUST have clear, verifiable, objective answers.
2. Avoid subjective questions.
3. Answers must come directly from the page content, do not make up or speculate.
4. Question type examples:
   (*@$\checkmark$@*) Good questions: "Who is the advisor of the student mentioned in the news?" (Requires linking Student Name from Page A to Advisor from Page B)
   (*@$\times$@*) Avoid: Questions answerable by a single page.
5. Answers should be concise and accurate.

(*@\textbf{Requirements:}@*)
1. The question MUST require integrating information from (*@\textbf{BOTH}@*) pages to answer.
2. The question should be related to the (*@\textcolor{blue}{\{domain\}}@*) domain.
3. Difficulty level: (*@\textcolor{blue}{\{difficulty\}}@*)
   - easy: Information is direct and obvious, easy to find
   - medium: Requires simple linking of facts across pages.
   - hard: Requires multi-hop reasoning and comparison.
4. Generate question and answer in English

Input Content:
(*@\textcolor{blue}{\{content\_page\_1\}}@*)
(*@\textcolor{blue}{\{content\_page\_2\}}@*)

(*@\textbf{[Output Format (Shared)]}@*)
Please return JSON format directly:
Please return JSON format directly:
{
    "question": "Question content",
    "answer": "Answer content"
}
\end{lstlisting}
\end{tcolorbox}

\begin{tcolorbox}[
enhanced jigsaw,
breakable,
pad at break*=1mm,
colback=white!95!gray,
colframe=gray!50!black,
title={Prompt for Teacher Model}]
\label{app:teacher model prompt}
\begin{lstlisting}[style=promptstyle]
You are a rigorous Quality Assurance Auditor and Content Editor for a QA dataset.
Your task is to perform a strict solvability check and refine the linguistic quality of the generated QA pair.

Input Data:
- Domain: (*@\textcolor{blue}{\{domain\}}@*)
- Source Content: (*@\textcolor{blue}{\{content\}}@*)
- Generated Question: (*@\textcolor{blue}{\{question\}}@*)
- Generated Answer: (*@\textcolor{blue}{\{answer\}}@*)

(*@\textbf{[Phase 1: Solvability Check (Gatekeeper)]}@*)
1. (*@\textbf{Strict Grounding}@*): Is the answer explicitly derivable from the Source Content? If it requires external knowledge or hallucination, REJECT immediately.
2. (*@\textbf{Logic Validation}@*): Does the answer directly address the question without logical gaps?

(*@\textbf{[Phase 2: Linguistic Refinement]}@*)
If the QA pair passes Phase 1, you must refine it:
1. (*@\textbf{Clarity}@*): Fix any ambiguous pronouns or vague entities (e.g., change "What is its release date?" to "What is the release date of [Entity]?").
2. (*@\textbf{Fluency}@*): Correct grammar and ensure the tone is professional and concise.
3. (*@\textbf{Objectivity}@*): Remove any subjective phrasing (e.g., "Why is it amazing?").

(*@\textbf{[Output Decision]}@*)
- If the Solvability Check fails -> Return "is_valid": false.
- If Solvability passes -> Return "is_valid": true AND provide the "refined_question" and "refined_answer".

Output Format (JSON):
{
    "is_valid": <true/false>,
    "refined_question": "The polished/corrected question (if valid)",
    "refined_answer": "The polished/corrected answer (if valid)",
    "reason": "Explanation of rejection or refinement details."
}
\end{lstlisting}
\end{tcolorbox}

\section{Error Category Case Studies}
\label{sec:error case}

To systematically diagnose the reliability of V-GEMS, we established a failure taxonomy derived from the agent cognitive pipeline. As autonomous web exploration requires a sequential execution of identifying, interpreting, and synthesizing information, we map our five error categories to breakdowns at specific stages of this cognitive process:

\begin{enumerate}
    \item \textbf{Navigational Robustness:} Can the agent physically locate the target context? (Category: \textit{Refusal})
    \item \textbf{Factual Grounding \& Reasoning:} Can the agent interpret the retrieved context faithfully without fabrication or logical fallacies? (Categories: \textit{Hallucination}, \textit{Totally Incorrect}. The difference between them show in ~\ref{tab:hallucination vs total})
    \item \textbf{Information Resolution:} Can the agent synthesize the answer with the required completeness and granularity? (Categories: \textit{Missing Key Info}, \textit{Imprecise})
\end{enumerate}

Below, we detail the characteristics of each error mode.

\subsection{Refusal}
Refusal occurs when the model explicitly explicitly declines to answer (e.g., outputs "I cannot find the information") or yields an empty response. While a refusal can be a valid response if the information truly does not exist, in our evaluation, this category specifically denotes \textbf{premature termination} due to navigational failures.

\subsection{Hallucination}
Hallucination represents a critical breakdown in \textit{factual grounding}. In these instances, the model confidently asserts specific information (e.g., dates, statistics, or named entities) that is objectively false or entirely absent from the retrieved DOM content.

This error mode characterizes the tension between an LLM's pre-trained parametric knowledge and the specific context of the webpage. Instead of extracting data from the source, the agent "overrides" the visual evidence with fabricated details, leading to answers that appear plausible but are extrinsic to the environment.

\subsection{Totally Incorrect}
Distinct from hallucination, Totally Incorrect errors occur when the agent attempts to reason over the correct source material but fails to derive the valid logical conclusion. Here, the predicted answer often contains entities actually present on the page, but they are misattributed or structurally flawed due to a \textit{logical mismatch}.

Common manifestations include conflating parallel entities (e.g., extracting the "Workshop Deadline" instead of the "Conference Deadline") or failing to satisfy negative constraints. This indicates that while the agent successfully located the data, it failed to process the structural relationships between elements.

\begin{table*}[ht]
    \centering
    \renewcommand{\arraystretch}{1.5} 
    \small
    
    \begin{tabularx}{\linewidth}{p{3.5cm} X}
        \toprule
        \textbf{Root Url} & \url{https://www.fiba.basketball/en} \\
        \midrule
        
        \textbf{Question} & Which team won the second place in the FIBA Asia Cup 2025? \\
        \midrule
        
        \textbf{Answer} & \textbf{China} \\
        \midrule
        
        \textbf{Source Website} & \url{https://www.fiba.basketball/en/history/195-fiba-asia-cup} \\
        \midrule
        
        \textbf{Website Information} & 
        \includegraphics[width=0.9\linewidth, keepaspectratio]{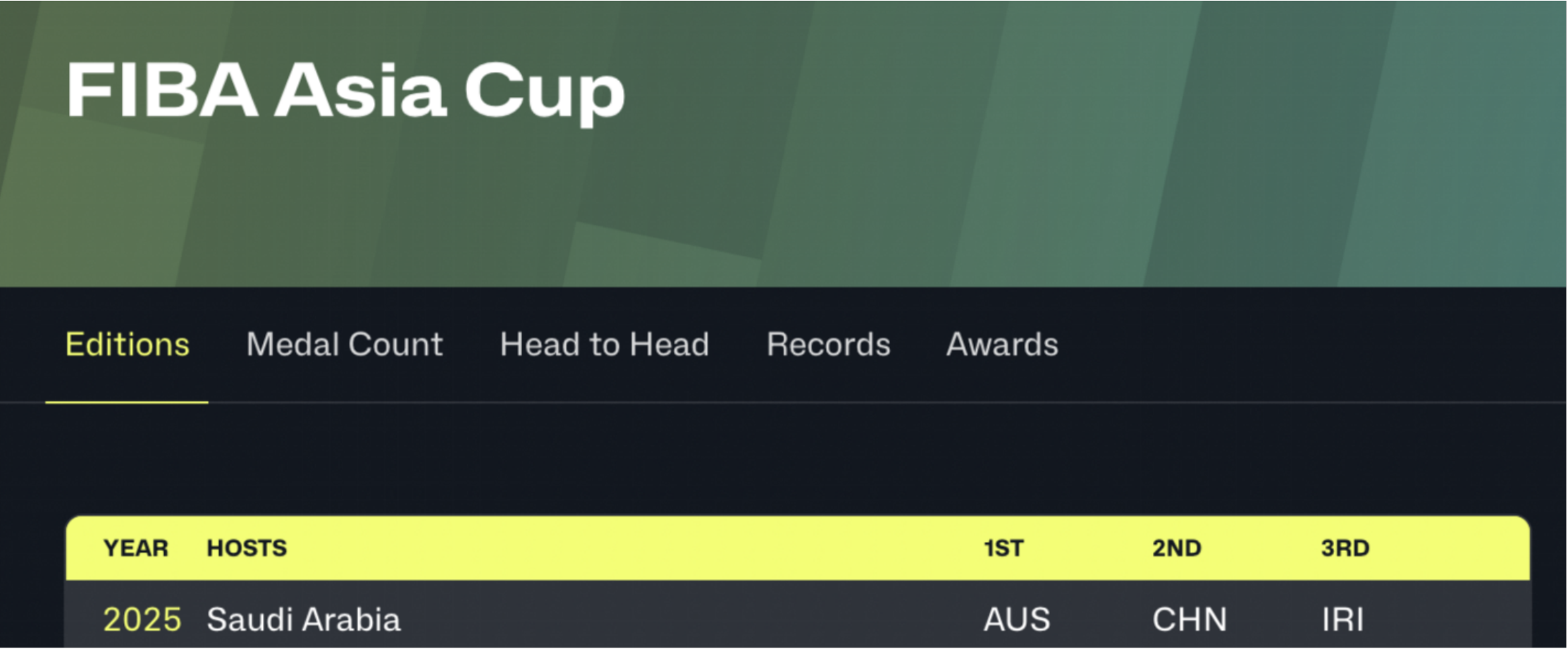} 
        \newline
        \textit{(Visual Context: The 2025 row shows 1st: AUS, 2nd: CHN, 3rd: IRI)} \\
        \midrule
        
        \textbf{Prediction \newline (Hallucination)} & 
        \textit{"Japan won the second place."} \newline
        \textcolor{hallucination_red}{\textbf{[Analysis]:}} The entity "Japan" is \textbf{fabricated}. It does not exist in the 2025 data row shown above. \\
        \midrule
        
        \textbf{Prediction \newline (Totally Incorrect)} & 
        \textit{"Australia won the second place."} \newline
        \textcolor{incorrect_orange}{\textbf{[Analysis]:}} "Australia" \textbf{exists} in the row (as 1st place) but is logically misattributed to the 2nd place. \\
        \bottomrule
    \end{tabularx}
    
    \caption{Case study comparing Hallucination and Totally Incorrect errors on the FIBA Asia Cup task.}
    \label{tab:hallucination vs total}
\end{table*}

\subsection{Missing Key Info}
As shown in ~\ref{tab:missing_key_info_fiba}, missing Key Info errors denote a failure in \textit{information synthesis} and \textit{recall}. In these cases, the model's response is factually accurate regarding the specific details it provides, but it fails to exhaustively cover the scope of the user's query.

Unlike hallucinations where the agent invents data, here the agent simply omits required components. For example, when asked for a comprehensive list or a multi-attribute description (e.g., "date and time"), the agent might return a truncated list or ignore specific constraints. This reflects a gap between successful page navigation and the ability to aggregate distributed information into a complete answer.

\subsection{Imprecise (Granularity Mismatch)}
\label{sec:imprecise_error}

\textbf{Imprecise} errors represent a failure in \textit{resolution alignment}. The predicted answer is correct in its general direction but lacks the specific granularity or fidelity required by the query.

As illustrated in Table \ref{tab:imprecise_example}, while the ground truth provides a precise timestamp ("21 Jul 2023"), the agent generalizes this to a broader scope ("Jul 2023"). Although this prediction is not factually contradictory, the loss of temporal fidelity fails to meet the analytical depth expected in automated web retrieval tasks. Such vague responses often render the output practically unusable for downstream applications that require exact timeline tracking.

\begin{table*}[h]
    \centering
    \renewcommand{\arraystretch}{1.5} 
    \small
    
    \begin{tabularx}{\linewidth}{p{3.5cm} X}
        \toprule
        \textbf{Question} & Which team won the FIBA Asia Cup 2025, and who was named the TISSOT MVP of the tournament? \\
        \midrule
        
        \textbf{Answer} & \textbf{Australia} won the FIBA Asia Cup 2025, and \textbf{Jaylin Galloway} was named the TISSOT MVP. \\
        \midrule
        
        \textbf{Prediction} & 
        \textit{"Australia won the FIBA Asia Cup 2025."} \newline
        \textcolor{missing_purple}{\textbf{[Analysis]:}} The answer is factually correct but \textbf{incomplete}. The agent correctly processed the first intent (Winning Team) but completely \textbf{ignored} the second intent (MVP) of the compound query, demonstrating a failure in comprehensive information recall. \\
        \bottomrule
    \end{tabularx}
    
    \caption{Case study of a Missing Key Info error on a multi-source query.}
    \label{tab:missing_key_info_fiba}
\end{table*}

\begin{table*}[t]
    \centering
    \renewcommand{\arraystretch}{1.5} 
    \small
    
    \begin{tabularx}{\linewidth}{p{3.5cm} X}
        \toprule
        \textbf{Question} & When was the last update made to the PSLE Secondary 1 Posting information? \\
        \midrule
        
        \textbf{Answer} & The last update to the PSLE Secondary 1 Posting information was made on \textbf{21 Jul 2023}. \\
        \midrule
        
        \textbf{Prediction \newline (Imprecise)} & 
        \textit{"The last update to the PSLE Secondary 1 Posting information was made in Jul 2023."} \newline
        \textcolor{imprecise_blue}{\textbf{[Analysis]:}} The answer is factually consistent with the truth but fails to meet the required \textbf{temporal granularity}. While the source provides a specific timestamp (Day-Month-Year), the model generalizes it to a broader scope (Month-Year), resulting in a loss of information fidelity. \\
        \bottomrule
    \end{tabularx}
    
    \caption{Case study of an Imprecise error. }
    \label{tab:imprecise_example}
\end{table*}

\end{document}